\documentclass[journal]{IEEEtran}
\usepackage{amsmath,amsfonts}
\usepackage{algorithmic}
\usepackage{algorithm}
\usepackage{array}
\usepackage[caption=false,font=normalsize,labelfont=sf,textfont=sf]{subfig}
\usepackage{textcomp}
\usepackage{stfloats}
\usepackage{url}
\usepackage{verbatim}
\usepackage{graphicx}
\usepackage{cite}

\newcommand{\mypara}[1]{\smallskip\textbf{#1}}
\usepackage{xcolor}
\usepackage{tabularray}
\usepackage{booktabs}  
\usepackage{multirow}
\usepackage{graphicx}  
\usepackage{amsmath}   

\begin{document}

\title{Amodal SAM: A Unified Amodal Segmentation \\ Framework with Generalization}

\author{Bo Zhang, Zhuotao Tian*, Xin Tao, Songlin Tang,  Jun Yu, Wenjie Pei*

\thanks{*Wenjie Pei and Zhuotao Tian are corresponding authors.}
\thanks{Bo Zhang, Zhuotao Tian, Jun Yu and Wenjie Pei are with the Department of Computer Science, Harbin Institute of Technology at Shenzhen, Shenzhen 518057, China.}
\thanks{Xin Tao and Songlin Tang are with Kuaishou Technology, Shenzhen, Guangdong, China}}

\markboth{ IEEE TRANSACTIONS ON IMAGE PROCESSING}%
{Shell \MakeLowercase{\textit{et al.}}: Bare Demo of IEEEtran.cls for IEEE Journals}
\maketitle

\begin{abstract}
Amodal segmentation is a challenging task that aims to predict the complete geometric shape of objects, including their occluded regions. Although existing methods primarily focus on amodal segmentation within the training domain, these approaches often lack the generalization capacity to extend effectively to novel object categories and unseen contexts.  This paper introduces Amodal SAM, a unified framework that leverages SAM (Segment Anything Model) for both amodal image and amodal video segmentation. Amodal SAM preserves the powerful generalization ability of SAM while extending its inherent capabilities to the amodal segmentation task. The improvements lie in three aspects: (1) a lightweight Spatial Completion Adapter that enables occluded region reconstruction, (2) a Target-Aware Occlusion Synthesis (TAOS) pipeline that addresses the scarcity of amodal annotations by generating diverse synthetic training data, and (3) novel learning objectives that enforce regional consistency and topological regularization. Extensive experiments demonstrate that Amodal SAM achieves state-of-the-art performance on standard benchmarks, while simultaneously exhibiting robust generalization to novel scenarios. We anticipate that this research will advance the field toward practical amodal segmentation systems capable of operating effectively in unconstrained real-world environments.
\end{abstract}


\section{Introduction}
\IEEEPARstart{T}{he} human visual system possesses the ability to interpolate unseen information, particularly through amodal perception, our capacity to mentally complete partially occluded objects. This innate ability has inspired a computer vision task called amodal segmentation, which aims to predict complete object shapes, including their hidden portions.

\begin{figure}
    \includegraphics[width=\linewidth]{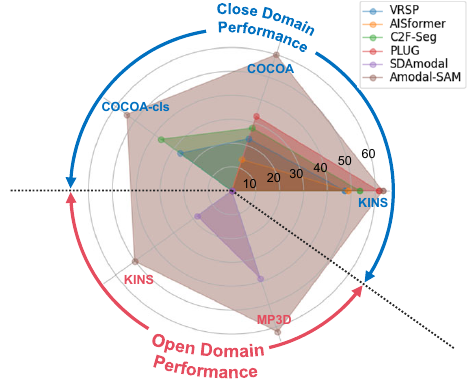}
    \caption{This figure shows the comparisons between our Amodal SAM and previous state-of-the-art models in the closed and open domains. In both settings, using $mIoU_{o}$ as the evaluation metric, our method outperforms the state-of-the-art model on all compared datasets.}
    \label{fig:teaser}
\end{figure}

Nonetheless, amodal segmentation remains challenging, as the existing methods primarily focus on amodal segmentation within the same domain witnessed during training. However, in open-world scenarios, models must generalize to novel object categories and contexts that have not been seen during model training, which poses challenges to current amodal segmentation models, as shown in Fig. \ref{fig:teaser}.

To enhance open-world visual perception capabilities, the Segment Anything Model (SAM)~\cite{kirillov2023segment} was introduced, demonstrating remarkable generalization to new samples. However, SAM is inherently constrained to segment visible regions, thereby lacking the capacity to address open-world challenges in amodal segmentation. This raises the question: \textit{Is it viable to effectively adapt the SAM model for amodal segmentation while preserving its generalization prowess?}

Therefore, in this work, we present Amodal SAM, a framework that extends SAM's capabilities to amodal segmentation while preserving its powerful zero-shot ability. Our key insight is that successful adaptation requires a holistic approach addressing three complementary aspects: model architecture, training data, and optimization objectives. Specifically, for model adaptation, we introduce a lightweight Spatial Completion Adapter (SCA) that enables the model to reconstruct occluded regions while maintaining SAM's core segmentation abilities. To overcome the scarcity of large-scale amodal annotations required by the open-world training, we propose Target-Aware Occlusion Synthesis (TAOS), an efficient pipeline that synthesizes diverse occlusion patterns from existing segmentation datasets without manual labeling. Finally, we introduce the learning objectives regarding regional consistency and holistic topological regularization to facilitate training. 

The effectiveness of the proposed Amodal SAM is demonstrated through extensive experiments across extensive datasets on both challenging image and video benchmarks. Amodal SAM not only achieves state-of-the-art performance on standard amodal segmentation tasks but also shows strong generalization to novel object categories and scenes. Furthermore, we show that our framework can be seamlessly extended to video amodal segmentation by adapting SAM-2, highlighting its flexibility and broad applicability. To the best of our knowledge, it is the first attempt to address video amodal segmentation in the open-world settings. 

To summarize, our contributions are as follows:
\begin{itemize}
    \item We observe that existing amodal segmentation models lack open-world 
    generalization capabilities, significantly limiting their real-world applications. 
    \item With the improvements across three dimensions - model, data, and 
    optimization - we successfully adapt SAM, a generic foundation model, to 
    amodal segmentation, allowing the model to generalize to novel and diverse 
    open-world scenarios.
    \item The obtained framework, \textit{i.e.}, Amodal SAM, achieves superior 
    performance in both closed and open-world scenarios, and can be easily 
    extended to video applications, marking the first successful implementation 
    of open-world video amodal segmentation.
\end{itemize}

\section{preliminaries}
\label{sec:prelimlaries}

\subsection{Amodal Object Segmentation}
\mypara{Task Description.}
Amodal object segmentation~\cite{zhu2016semanticamodalsegmentation,li2016amodalinstancesegmentation,nguyen2021weaklysupervisedamodalsegmenter,tran2024shapeformershapepriorvisibletoamodal,tran2024aisformeramodalinstancesegmentation,qi2019amodal,zhan2024amodalgroundtruthcompletion}addresses the challenge of inferring the integral geometry of an object, encompassing both its visible regions and the regions obscured behind occluders.Formally, according to~\cite{qi2019amodal,tran2024aisformeramodalinstancesegmentation,tran2024shapeformershapepriorvisibletoamodal}, a rough bounding box $B$ that covers the region-of-interest (ROI) in a specific image $I$ will be provided, to identify both the visible and occluded regions of the target object through the prediction mask $M$. With the amodal segmentation model $\mathcal F$, this procedure can be formulated as:
\begin{equation}
    \label{eqn:amodal_seg}
    M = \mathcal{F}(B, I).
\end{equation}
Certain approaches~\cite{qi2019amodal,li2016amodalinstancesegmentation,follmann2018learninginvisibleendtoendtrainable,Zhang_2019} may use conventional segmentation algorithms, such as~\cite{ronneberger2015unet,long2015fullyconvolutionalnetworkssemantic,he2017mask}, to explicitly derive the visible region mask from the region-of-interest (ROI) box $B$ corresponding to the target object as an input component. In this case, the formulation becomes:
\begin{equation}
    \label{eqn:amodal_seg_alternative}
    M = \mathcal{F}(s(B), I),
\end{equation}

where $s()$ denotes the conventional segmentation algorithm that transforms the visible part within the ROI box $B$ into a mask. For both Eqs. \ref{eqn:amodal_seg} and \ref{eqn:amodal_seg_alternative}, the ground truth mask of the target object is provided to supervise the predicted mask $M$ during training. 

\mypara{Challenges in Open-World Scenarios.}
Despite significant advancements, conventional amodal segmentation methodologies predominantly operate under a closed-world assumption, restricted to a predefined taxonomy of object categories~\cite{zhu2016semanticamodalsegmentation,qi2019amodal,follmann2018learninginvisibleendtoendtrainable}. Such a paradigm inherently lacks the flexibility to generalize to novel or "out-of-distribution" (OOD) objects encountered in complex, unstructured environments. Furthermore, evidence suggests that existing models exhibit a pronounced sensitivity to domain shifts; for instance, representations learned from curated indoor datasets often undergo substantial performance degradation when deployed in disparate contexts, such as autonomous driving or aerial surveillance.

This decrease in robustness can be attributed to the multifaceted variations in object appearances, diverse textures, and the intricate, often unpredictable, occlusion patterns characteristic of real-world scenes. Consequently, there is an exigent need for a category-agnostic amodal segmentation framework capable of zero-shot generalization. In this work, we mitigate these limitations by harnessing the potent zero-shot capabilities and rich geometric priors of the Segment Anything Model (SAM)~\cite{kirillov2023segment}. By bridging the gap between SAM's promptable modal segmentation and the requirement for amodal reasoning, we aim to achieve a more resilient and versatile solution for open-world perception.

\begin{figure*}[t]
    \centering
    \includegraphics[width=\linewidth]{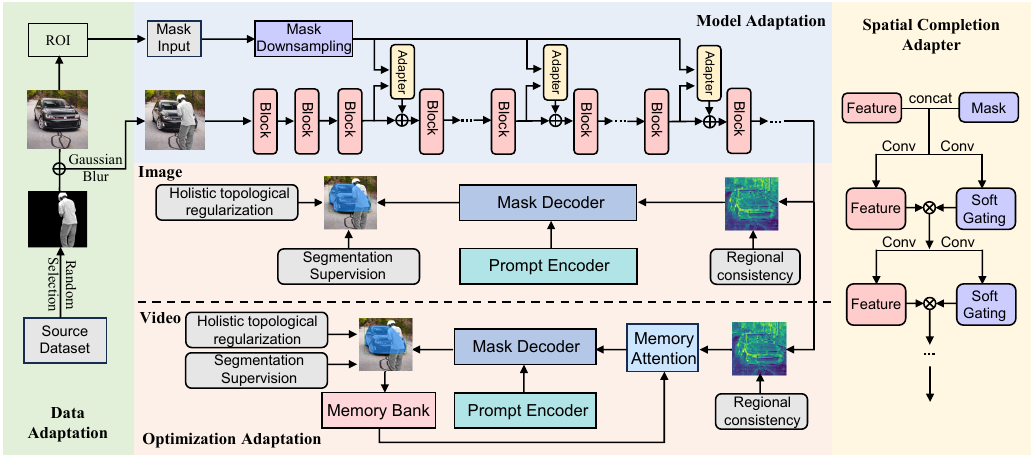}
\caption{The overall structure of the proposed Amodal Segment Anything Model(SAM) includes three key aspects for the adaptation from SAM to Amodal SAM: data, model, and optimization. The process of data adaptation involves an automated pipeline for converting generic segmentation annotations into those necessary for amodal segmentation. Model adaptation is realized through the Spatial Completion Adapter (SCA) in a manner focused on the encoder-tuning. Optimization adaptation is performed using a composite learning objective.}
    \label{fig:architecture}
    
\end{figure*}

\subsection{Segment Anything Model (SAM)}

\mypara{Architectural Overview.} The Segment Anything Model (SAM)~\cite{kirillov2023segment}, pioneered by Meta AI, serves as a versatile foundation model for promptable image segmentation. Its architecture comprises a tripartite design: 1) a heavyweight Vision Transformer (ViT)-based image encoder that maps high-resolution inputs into dense visual embeddings; 2) a flexible prompt encoder that represents sparse queries (e.g., points, boxes, text) or dense masks as positional encodings; and 3) a lightweight mask decoder that integrates these multimodal signals through cross-attention to generate precise segmentation masks. Trained on the colossal SA-1B dataset—encompassing 11 million images and over 1.1 billion masks—SAM demonstrates extraordinary zero-shot generalization across diverse visual domains, effectively establishing a new paradigm for interactive and automated segmentation.

\mypara{SAM-based adaptation.} The remarkable generalization capability of SAM has inspired numerous downstream adaptation~\cite{chen2023samfailssegmentanything,li2023semanticsamsegmentrecognizegranularity,MedSAM,ke2023segmenthighquality} tailored for specialized domains. Notable advancements include HQ-SAM~\cite{ke2023segmenthighquality}, which introduces a refined output token to capture intricate boundary details, and MedSAM~\cite{MedSAM}, which specializes the model for medical imaging via large-scale domain-specific fine-tuning. Drawing inspiration from the adapter-based paradigm in natural language processing~\cite{houlsby2019parameterefficienttransferlearningnlp,hu2021loralowrankadaptationlarge}, recent frameworks such as SAM-Adapter~\cite{chen2023samfailssegmentanything} and ViT-Adapter~\cite{chen2023visiontransformeradapterdense} utilize lightweight bottleneck modules to inject domain-specific knowledge into the frozen ViT backbone. These strategies preserve the robust pre-trained representations while enabling efficient knowledge transfer to downstream tasks with minimal computational overhead.

\mypara{Limitations in amodal segmentation.} Despite the success of these adaptation schemes, extending SAM’s capabilities to the amodal domain presents a unique challenge. While we adopt the principle of adapter-based tuning, we observe that the naive adapter-based adaptation schemes yield suboptimal performance for this challenging task as shown in later experiments. This inadequacy stems from a fundamental representational gap: SAM is inherently "modal-centric," having been optimized to delineate visible object boundaries. Amodal segmentation, conversely, requires occlusion-aware hallucination and geometric reasoning to infer the morphology of regions sequestered behind occluders. Simple adapters often fail to bridge this gap as they lack the specialized mechanisms needed to disentangle visible features from occluded priors. In the following sections, we demonstrate that a more nuanced architectural modification is requisite to unlock SAM's potential for amodal segmentation.

\section{Amodal SAM}

\subsection{Overview}

As amodal segmentation in the open world requires robust zero-shot generalization to handle novel object classes and data distributions, in this work, we propose Amodal SAM - a framework that decently extends SAM's capabilities from visible region segmentation to amodal segmentation while preserving SAM's strong zero-shot capabilities. The proposed paradigm focuses on the training phase of the following three synergistic aspects: 

\begin{itemize}
    \item \textit{Model adaptation} enables efficient adaptation with minimal structural modification through the lightweight gated adapter.
    \item \textit{Data adaptation} facilitates the supervised tuning via synthesizing object occlusions without the need for exhaustive human annotation.
    \item \textit{Optimization adaptation} involves transitioning from a generic vision foundation model to a specialized amodal segmentation model by incorporating specific learning objectives.
\end{itemize}

In the following Sections, Section \ref{sec:model_adaptation}, Section \ref{sec:data_adaptation} and Section \ref{sec:optimization_adaptation} will detail our adaptation strategies for model architecture, training data, and optimization processes, respectively. Subsequently, Section \ref{sec:video amodal segmentation} demonstrates how our designs can be easily extended to video amodal segmentation with SAM-2\cite{ravi2024sam2segmentimages}, highlighting the generalization capabilities.

\subsection{Model Adaptation}
\label{sec:model_adaptation}
In this study, we investigate adapters for the model adaptation to enhance the capacity in occluded region perception. 

\mypara{Encoder-focused adaptation.}
Instead of inserting the adapters into both the encoder and decoder of SAM, we adopt an encoder-focused adaptation strategy.

Specifically, the SAM encoder handles feature extraction, while the decoder primarily converts features into masks. Consequently, the encoder may encounter a more substantial domain disparity when facing different tasks. If decoder tuning were implemented, it might jeopardize SAM's inherent mask generation abilities, as later demonstrated in \ref{sec:ablation_study}. Therefore, an encoder-focused adaptation can bridge the domain gap across diverse tasks while preserving the model's core functionality with the frozen decoder. The overall modeling process of Amodal SAM can be expressed as:

\begin{equation}
\label{eqn:enc_first}
    M = \mathcal{F}_{\text{dec}}(\mathcal{F}_{\text{prompt}}(B), \mathcal{F}'_{\text{a-enc}}(I)),
\end{equation}
where $M$ is the predicted mask for both visible and occluded regions of the target object, which is specified by an ROI box $B$ on the input image $I$. $\mathcal{F}_{\text{dec}}$ and $\mathcal{F}_{\text{prompt}}$ denote SAM's mask decoder and prompt encoder, respectively, while $\mathcal{F}'_{\text{a-enc}}$ represents the adapted SAM encoder tailored for amodal segmentation.

\mypara{Prior-guided feature extraction.}
For the amodal segmentation task, an ROI box $B$ is needed in the input to indicate the target. We found it is beneficial to additionally incorporate a binary prior mask $M_\text{spec}$ into the feature encoder, as the spatial prior to guide the target-related feature generation. During inference, $M_\text{spec}$ can be obtained from $B$ by setting the value to 1 for the regions inside $B$ and to 0 for the remaining regions of the binary mask $M_\text{spec}$. To this end, Eq.~\eqref{eqn:enc_first} is accordingly updated as: 
\begin{equation}
\label{eqn:enc}
    M = \mathcal{F}_{\text{dec}}(\mathcal{F}_{\text{prompt}}(B), \mathcal{F}'_{\text{a-enc}}(I,M_\text{spec})).
\end{equation}

It is important to note that the mask $M_\text{spec}$ might not be accessible during model training, given the absence of a dataset containing both occluded and full segmentation ground-truth masks of the target object. Our approach to tackle this challenge will be presented in \ref{sec:data_adaptation}.

\begin{figure}[]
    \centering
    \includegraphics[width=\linewidth]{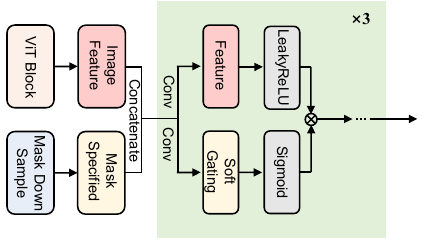}
\caption{The Spatial Completion Adapter combines the image features and $M_{spec}$ by concatenating them as input, and then iterates the feature selection and completion process three times.}
    \label{fig:sca}
\end{figure}

\mypara{Spatial Completion Adapter (SCA).} In Eq. \ref{eqn:enc}, incorporating the spatial prior via $M_\text{spec}$ into the encoder underscores the need to devise a dedicated encoder capable of accommodating this guided enhancement. Hence, we propose the Spatial Completion Adapter (SCA) and integrate it into the baseline SAM encoder. SCA is designed to reconstruct obscured regions of target objects within the feature space, leveraging the spatial cues offered by $M_\text{spec}$, thereby facilitating the completion of occluded areas.

Unlike conventional adapters~\cite{li2020unsuperviseddeepmetriclearning,yuan2021tokenstotokenvittrainingvision,ranftl2021visiontransformersdenseprediction} that rely on basic convolutions or linear transformations for cross-domain feature adaptation, the proposed SCA draws inspiration from Gated Convolution~\cite{Yu_2019_ICCV}, implements a dynamic feature selection mechanism. As illustrated in Fig. \ref{fig:sca}, the SCA first concatenates the latent features $\mathbf{E} \in \mathbb{R}^{W \times H \times C}$ from the preceding ViT block with the spatially downsampled mask $M_\text{spec}$. This composite representation is then processed through two parallel convolutional branches to derive element-wise gating weights $\mathbf{G}$ and transformed features, respectively. Formally, this spatially-informed modulation is expressed as:

\begin{equation}
\label{eqn:acadapter}
\begin{split}
    & \mathbf{G} = \sigma(\mathcal{F}_\text{gate}(\mathbf{E}, M_\text{spec})), \\
    & \mathbf{O} = \mathbf{G} \odot \phi(\mathcal{F}_\text{feat}(\mathbf{E},M_\text{spec})).
\end{split}
\end{equation}

\begin{figure}[t]
    \includegraphics[width=\linewidth]{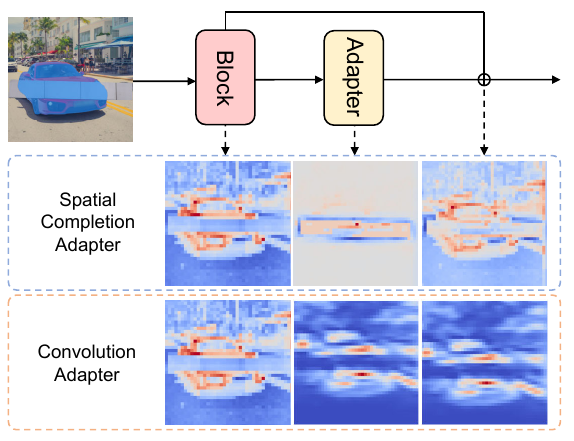}
    \caption{
    The figure demonstrates the effects of the Spatial Completion Adapter (SCA). It is evident that SCA effectively complements the input by restoring features in the occluded regions.}
    \label{fig:gating}
\end{figure}

Here, $\mathbf{G}\in \mathbb{R}^{W\times H\times C}$ and $\mathbf{O}\in \mathbb{R}^{W\times H\times C}$ represent the learned gating weights and filtered features, respectively. $\mathcal{F}_\text{gate}$ and $\mathcal{F}_\text{feat}$ are implemented via vanilla convolutional layers, while $\sigma$ (Sigmoid) and $\phi$ (LeakyReLU~\cite{_enyi_it_2014}) serve as activation functions. By constraining the gating weights $\mathbf{G}$ within $[0, 1]$, the SCA functions as a soft switching mechanism that prioritizes information recovery in regions highlighted by the spatial prior (see Fig. \ref{fig:gating}).

In our implementation, this feature-completion operation is executed three times within each SCA module to ensure sufficient representational capacity. We strategically integrate the SCAs into the shallow, middle, and deep layers of the SAM image encoder. The empirical efficacy of this hierarchical integration is further validated in Section \ref{sec:ablation_study}.

\begin{figure*}[t]
    \centering
    \includegraphics[width=0.9\linewidth]{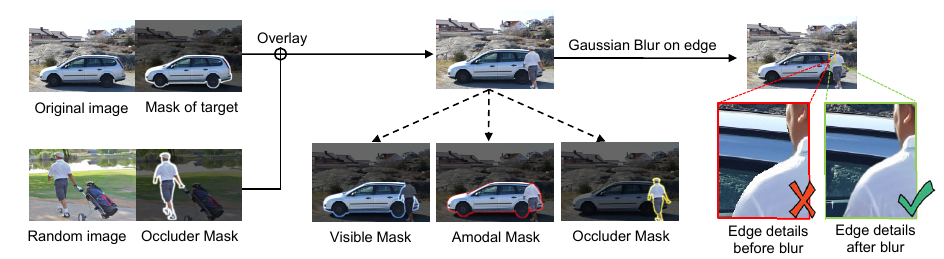}
    \caption{The illustration of the proposed Target-Aware Occlusion Synthesis (TAOS) pipeline. Initially, by randomly selecting masks and superimposing them on the target object, we create an amodal mask, a visible mask, and an occlusion mask. Furthermore, applying a Gaussian blur to the edges helps in effectively smoothing the transition between the two objects. Details can be found in Section~\ref{sec:data_adaptation}.}
    \label{fig:dataconstruct}
\end{figure*}

\subsection{Data Adaptation}
\label{sec:data_adaptation}

Amodal segmentation poses a fundamental challenge that sets it apart from traditional object segmentation by necessitating the precise delineation of obscured object regions. 

However, as the data scale is essential to achieve an open-world model, existing amodal segmentation datasets~\cite{zhu2016semanticamodalsegmentation,follmann2018learninginvisibleendtoendtrainable,qi2019amodal} are limited in scale and scope, typically covering only specific domains with restricted object categories, making them inadequate for open-world applications. Moreover, manually curating a large-scale dataset with sufficient amodal masks for occluded objects would be prohibitively time-consuming and expensive.

To address these challenges, we introduce a Target-Aware Occlusion Synthesis (TAOS) pipeline to accomplish the data adaptation. The TAOS pipeline can efficiently convert standard segmentation annotations available in the large-scale SA-1B dataset~\cite{kirillov2023segment} into the required formats for amodal training, eliminating the necessity for manual labeling. Further details are outlined below.

\mypara{Target-Aware Occlusion Synthesis (TAOS).} Since the current extensive segmentation dataset only labels the visible regions, within TAOS, we choose to artificially create and model occlusions that could potentially occur among distinct objects.

To generate an image featuring an occluded object, we first randomly select an image from the original dataset, \textit{i.e.}, SA-1B, containing an object within a predefined size range, designating this object as the \textit{target}. Subsequently, we randomly crop an object (or a portion thereof) from another randomly selected image, ensuring that it is of a suitable size relative to the target object, to serve as the \textit{occluder}. Then, we overlay the \textit{occluder} onto the \textit{target} object at a random position and with a random overlapping area within a specified range. Furthermore, we employ VLM to evaluate the generated occlusion and eliminate invalid data.

Finally, we apply pixel blurring to the boundaries of the occluded regions to ensure the naturally synthesized occlusion. Specifically, we adopt Gaussian Blur, which smooths each boundary pixel by normalizing the pixel value according to a predefined Gaussian kernel. Formally, a boundary pixel located at $(x, y)$ in an image $I$ is smoothed as follows: 

\begin{equation}
\begin{split}
    G(x, y) &= \frac{1}{2 \pi \sigma^2} \exp\left(-\frac{x^2 + y^2}{2 \sigma^2}\right), \\
    I'(i, j) &= \sum_{x=-\frac{k}{2}}^{\frac{k}{2}} \sum_{y=-\frac{k}{2}}^{\frac{k}{2}} G(x, y) \cdot I(i + x, j + y). 
\end{split}
\end{equation}

Here, $G$ denotes the Gaussian kernel with size $k$. The overall pipeline is illustrated in Fig. \ref{fig:dataconstruct}.

\begin{figure*}[]
    \centering
    \includegraphics[width=0.9\linewidth]{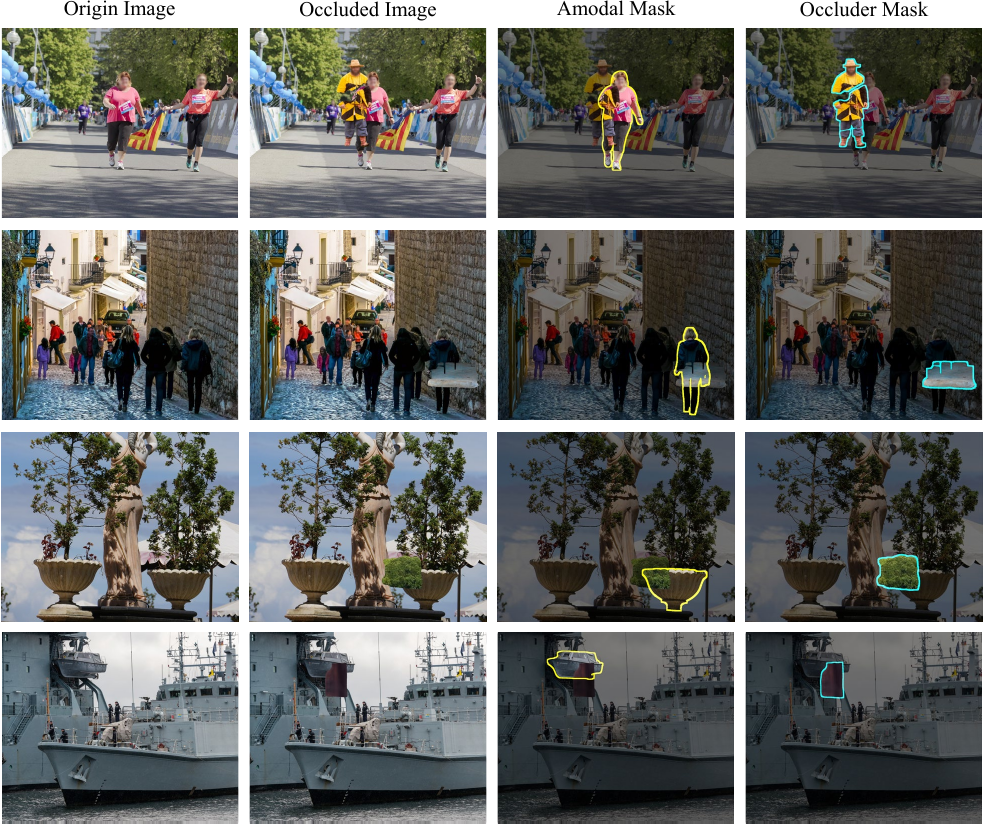}
    \caption{Examples of the datasets we constructed. For each example showing the original image, the image after the occlusion was added, the amodal mask, and the mask of the occluder, respectively.}
    \label{fig:visible_data}
\end{figure*}

\subsection{Optimization Adaptation}
\label{sec:optimization_adaptation}
The model architecture and training data adaptations necessitate refining SAM's optimization process to better suit the amodal segmentation requirements.In addition to the vanilla segmentation loss $\mathcal{L}_\text{seg}$, we introduce two additional learning objectives for amodal segmentation: (1) regional consistency $\mathcal{L}_\text{reg}$ between the visible and occluded regions of the target object, and (2) holistic topological regularization $\mathcal{L}_\text{hol}$ based on adversarial learning.  To this end, the overall learning objective $\mathcal{L}$ is formulated as:

\begin{equation}
    \mathcal{L} =  \mathcal{L}_\text{seg} + \mathcal{L}_\text{reg} +  \mathcal{L}_\text{hol}.
\end{equation}

\mypara{Segmentation Supervision.}To begin with, we first adopt Dice loss and BCE loss to optimize the segmentation prediction:

\begin{equation}
    \mathcal{L}_\text{seg} = \mathcal{L}_\text{Dice} + \alpha \mathcal{L}_\text{BCE},
\end{equation}
Where $\alpha$ is a balancing factor that is set to 10, following 
~\cite{cheng2021perpixelclassificationneedsemantic,MedSAM,
chen2023samfailssegmentanything}.

\mypara{Regional consistency.}
Then, based on the rationale that visible and occluded regions of the same objects are expected to exhibit similar characteristics, such as appearance and texture patterns, we incorporate intra-object regional consistency during training. This is achieved by minimizing the difference between the representations of visible and occluded regions:

\begin{equation}
\centering
\begin{split}
    &\mathbf{E}_\text{vis} = \frac{\sum \mathbf{E} \odot \mathbf{M}_\text{vis}}{\sum \mathbf{M}_\text{vis}}, \quad \mathbf{E}_\text{occ} = \frac{\sum \mathbf{E} \odot \mathbf{M}_\text{occ}}{\sum \mathbf{M}_\text{occ}},\\
    &\mathcal{L}_\text{reg} = 1 - \texttt{cos}(\mathbf{E}_\text{vis}, \mathbf{E}_\text{occ}).
\end{split}
\end{equation}

Here, $\mathbf{E}\in \mathbb{R}^{W\times H\times C}$ denotes the feature map extracted by the adapted encoder $\mathcal{F}'_{\text{a-enc}}(I,M_\text{spec})$. With the visible mask $\mathbf{M}_\text{vis} = 1 - \mathbf{M}_\text{occ}$ and the occlusion mask $\mathbf{M}_\text{occ}$, we can obtain $\mathbf{E}_\text{vis}\in \mathbb{R}^{C}$ and $\mathbf{E}_\text{occ}\in \mathbb{R}^{C}$, indicating the representations of the visible and occluded regions, respectively. The function $\texttt{cos}()$ calculates the cosine similarity between the inputs. Therefore, by minimizing the regional consistency loss $\mathcal{L}_\text{reg}$, the visible and occluded regions are encouraged to have more similar representations.

\mypara{Holistic topological regularization.}Furthermore, we introduce a topological regularization that preserves the high-order structural relationships between the predicted mask and ground truth, enabling the explicit learning of shape priors through adversarial training. Specifically, we implement a discriminator $D$ that evaluates the topological and morphological consistency between the predicted mask and ground truth. The optimization objective for the discriminator $D$ is formulated as:

\begin{equation}
\begin{split}
    \mathcal{L}_\text{hol} = &\min_\text{A-SAM}\max_D (\mathbb{E}_{M_g \sim P_\text{GT}(M)}[\log D(M_g, I)] \\
    &+ \mathbb{E}_{M_p \sim P_\text{A-SAM}(M)}[\log (1-D(M_p, I))]),
\end{split}
\end{equation}

where $P_\text{GT}$ and $P_\text{A-SAM}$ are the distributions of ground truth masks and predicted masks by the proposed Amodal SAM, respectively. 

To effectively extend the capabilities of SAM from the visible region segmentation to amodal segmentation while preserving its powerful zero-shot functionality, we propose Optimization Adaptation, as detailed in Section 3.4 of the main paper. This approach includes holistic topology regularization, which is based on adversarial learning\cite{gan} and necessitates the employment of a discriminator for implementation.

\begin{figure}[]
    \centering
    \includegraphics[width=\linewidth]{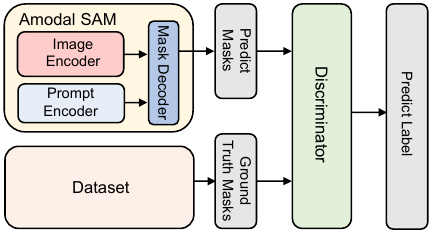}
    \caption{The discriminator uses the predictions from Amodal SAM as negative samples and the ground truth as positive samples. This alignment helps the model-predicted mask distribution closely match the ground truth, facilitating the learning of shape priors for achieving topological regularization.}
    \label{fig:discriminator}
\end{figure}

Our Target-Aware Occlusion Synthesis (TAOS) technique enables the generation of occluded images while retaining the original unoccluded image. In the training of the open-domain model, the discriminators receive the original image and its corresponding mask as input. Specifically, the images and masks are concatenated, and their features are extracted using a multilayer network. These features undergo progressive downsampling, and the alignment between the predicted masks and the ground truth in terms of topology and morphology is assessed. To ensure robust consistency measurements, a sigmoid activation function is applied to confine the output values within a probability range of 0 to 1. During closed-domain training, where unobstructed images are not available, only the mask is utilized as input to the discriminator. In both training phases, the output from Amodal SAM serves as negative samples, while the ground truth acts as positive samples. This approach aims to align the distribution of the model-predicted mask with the ground truth, facilitating the learning of shape priors for achieving topological regularization as illustrated in Fig. \ref{fig:discriminator}.

\begin{figure}[t]
    \centering
    \includegraphics[width=1\linewidth]{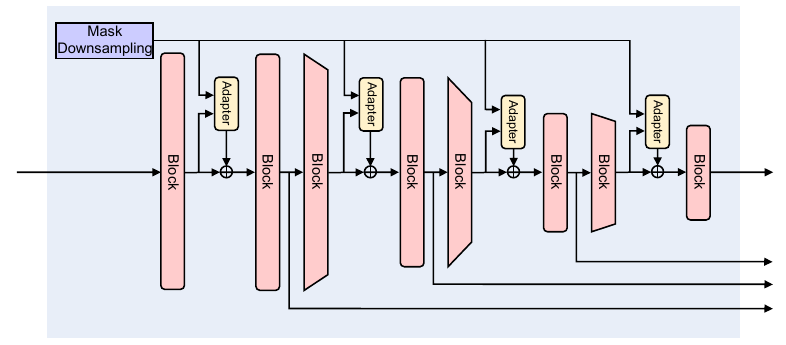}
    \caption{The image encoder of SAM-2 consists of four stages with distinct spatial scales. We incorporate an SCA at the outset of each stage to ensure that every feature output from the image encoder undergoes spatial completion.}
    \label{fig:sam2}
\end{figure}

\subsection{Video Amodal Segmentation}\label{sec:video amodal segmentation}
\mypara{From SAM to SAM-2: Architectural Evolution.} SAM-2~\cite{ravi2024sam2segmentimages} represents a significant paradigm shift, extending the foundation model's efficacy from static images to the temporal domain. While retaining the core tripartite structure of its predecessor—comprising an image encoder, prompt encoder, and mask decoder—SAM-2 introduces a memory bank and a memory attention layer. These modules facilitate the fusion of current frame features with historical memory embeddings, enabling robust spatio-temporal context integration. A key design transition is the replacement of the vanilla ViT with a hierarchical Hiera encoder~\cite{ryali2023hierahierarchicalvisiontransformer}, which extracts multiscale features across four stages. This structural consistency provides a seamless pathway for extending our amodal adaptation strategies to video sequences.

\mypara{Hierarchical SCA Integration.} To accommodate the multiscale nature of the Hiera-based encoder, we distribute the Spatial Completion Adapter (SCA) across each of the four encoding stages. Early-stage SCAs enhance high-resolution features critical for boundary refinement, while late-stage SCAs modulate the coarse embeddings that guide the final mask generation. By injecting the SCA into each stage, we effectively leverage SAM-2's hierarchical representation to perform comprehensive amodal segmentation across disparate spatial scales, as illustrated in Fig. \ref{fig:sam2}.

\mypara{Progressive Temporal Propagation.} Transitioning from amodal segmentation to video presents unique challenges due to the temporal volatility of occlusion patterns and the dynamic trajectories of objects. In this context, the frame-by-frame derivation of the spatial prior $M_\text{spec}$ is often insufficient. To resolve this, we propose a progressive temporal approach that exploits spatio-temporal coherence. Specifically, the prediction output from the preceding frame is utilized to generate a coarse Region of Interest (ROI), which subsequently serves as the $M_\text{spec}$ input for the current frame. This recursive propagation mechanism ensures that the model maintains a persistent "amodal memory" of the object's position, even under severe or transient occlusions.

\mypara{Optimization and Training.} Following the original SAM-2 training protocol, we employ a joint image-video training strategy. To further fortify the amodal reasoning capability, we incorporate the optimization objectives from our image-based framework, including segmentation supervision, region consistency, and holistic topological regularization. This synergistic integration not only validates the generalization of our SCA design but also establishes a resilient framework for open-world amodal video segmentation.

\section{Experiments}
\label{sec:experiments}
\subsection{Experimental Setup}
\mypara{Data sources of TAOS pipeline.}The SA-1B dataset~\cite{kirillov2023segment} comprises 11 million varied, high-resolution, privacy-protecting images and 1.1 billion top-notch segmentation masks obtained through SAM's data engine. This dataset offers well-labeled, category-independent, high-quality, multi-granularity object masks. Utilizing these labeled masks enables the acquisition of necessary amodal annotations without human intervention via the proposed TAOS pipeline, as elaborated in Section \ref{sec:data_adaptation}.

\mypara{Amodal segmentation benchmarks.}To assess the effectiveness of our approach, we evaluate Amodal SAM on six amodal segmentation datasets including both image and video amodal segmentation benchmarks: KINS\cite{qi2019amodal}, COCOA\cite{zhu2016semanticamodalsegmentation}, COCOA-cls\cite{follmann2018learninginvisibleendtoendtrainable}, D2SA\cite{d2sa}, FISHBOWL\cite{fish}, and MOViD-A\cite{gao2023coarsetofineamodalsegmentationshape}.

\begin{itemize}
    \item \textbf{KINS}, an extensive traffic amodal dataset based on KITTI\cite{_enyi_it_2014}, comprises 14,991 images across 7 categories, with 7,474 images for training and the rest for testing. 
    \item \textbf{COCOA}, a subset of the COCO dataset\cite{lin2015microsoftcococommonobjects}, consists of 2,476 training images and 1,223 testing images.
    \item \textbf{COCOA-cls} is a dataset of objects selected from COCOA, containing 80 categories.
    \item \textbf{D2SA} is a supermarket product dataset based on D2S, designed to capture the environments and requirements of industrial applications.
    \item \textbf{FISHBOWL} is a video benchmark, captured from a publicly available WebGL demo of an aquarium.
    \item \textbf{MOViD-A} is a video-based synthesized dataset created from the MOViD dataset.
\end{itemize}

\mypara{Evaluation protocols.}For evaluation, we employ the mean Intersection over Union (mIoU) as the primary metric, following~\cite{gao2023coarsetofineamodalsegmentationshape,zhan2024amodalgroundtruthcompletion}. We compute both the ground-truth amodal mask (mIoU$_{f}$) and the occluded region (mIoU$_{o}$). The occluded mIoU evaluates the overall quality of the occluded section of the target object. mIoU$_{full}$ calculates the average Intersection over Union (IoU) between the predicted amodal masks (representing the entire object) and the ground-truth amodal masks. In contrast, mIoU$_{occ}$ measures the average IoU between the predicted occlusion masks (representing the occluded regions) and the ground-truth occlusion masks. It can be expressed as:

\begin{equation}
\label{eqn:miou}
\begin{split}
    & I_{occ} = (M_\text{pred} - M_\text{visible})\quad  \& \quad (M_\text{gt} - M_\text{visible}) \\
    & U_{occ} = (M_\text{pred} - M_\text{visible})\quad  | \quad (M_\text{gt} - M_\text{visible}) \\
    &  mIoU_{occ} = \frac{1}{n} \sum_{i=1}^{n} I_{occ} / U_{occ}
\end{split}
\end{equation}

\begin{figure}[]
    \centering
    \includegraphics[width=\linewidth]{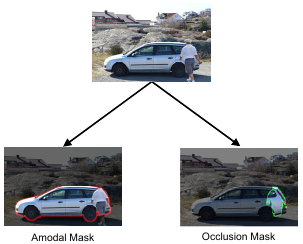}
    \caption{The occlusion mask delineates the occluded region within the amodal mask. 
    mIoU$_{f}$ assesses the predictive accuracy of the amodal mask, whereas mIoU$_{o}$ 
    evaluates the predictive accuracy of the occlusion mask. During evaluation, even if 
    the model's predicted mask solely includes visible regions, mIoU$_{f}$ might 
    surpass 50, whereas mIoU$_{o}$ could remain at 0.}
    \label{fig:miou}
\end{figure}

Where $M_\text{pred}$ denotes the predicted mask, $M_\text{gt}$ represents the ground truth mask, and $M_\text{visible}$ is the mask of the visible region. Fig. \ref{fig:miou} demonstrates the difference between amodal masks and occlusion masks. The mIoU$_{occ}$ metric specifically assesses the model's performance in predicting occluded regions, which is inherently more challenging than predicting visible regions in amodal segmentation. Therefore, mIoU$_{occ}$ is a crucial metric for evaluating model performance in this scenario.

\mypara{Implementation details.}We implement our method with PyTorch. In our experiments, we augment the bounding box with a mask as the model input, expanding the bounding box by a factor of 0.2 in all directions, resulting in an input area 1.44 times larger than the original. The iteration counts are set to 40k, 20k, 7k, 75k and 75k for KINS, COCOA, COCOA-cls, FISHBOWL and MOViD-A datasets, respectively. We utilize the AdamW optimizer with a learning rate that decreases from 1e-4 to 5e-5 throughout training.

\begin{figure*}[t]
    \centering
    \includegraphics[width=\linewidth]{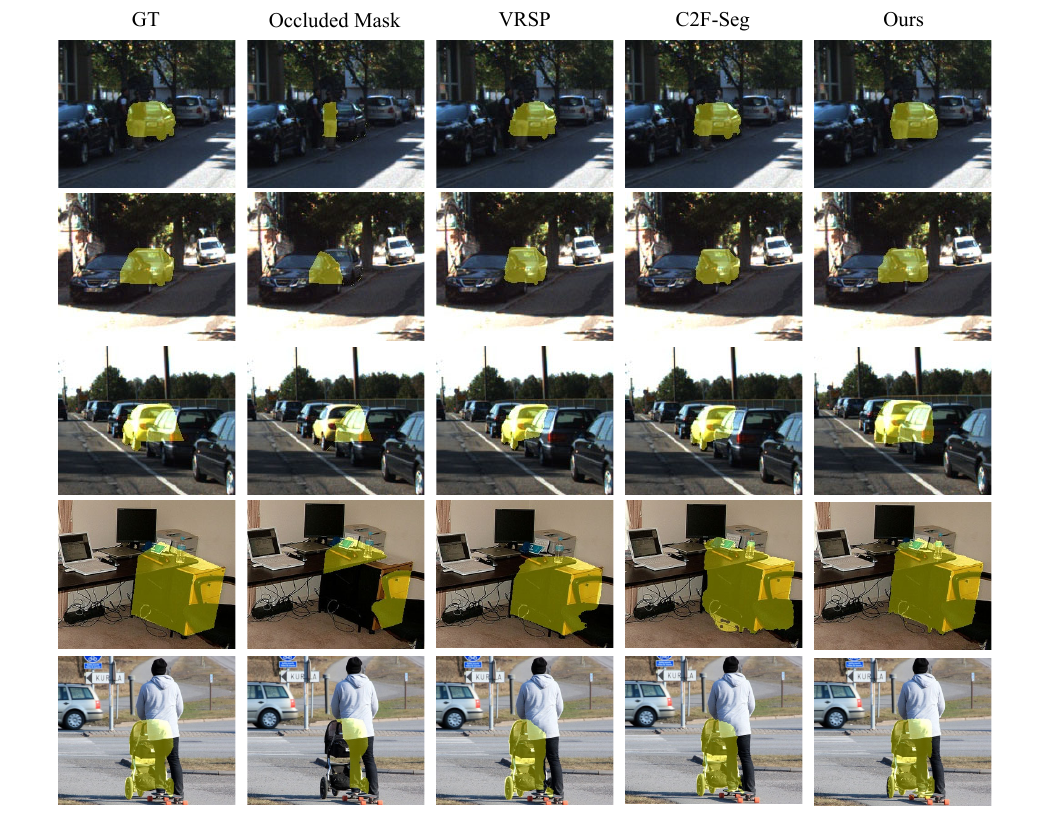}
    \caption{This figure presents the qualitative results of VRSP, C2F-Seg, 
    and our method on the KINS and COCOA datasets. ``GT" represents the ground 
    truth, and the ``Occluded mask" indicates the occluded region of the target 
    object.}
    \label{fig:visualization1}
\end{figure*}

\begin{table}
  \caption{Performance comparison on three representative amodal datasets.}
  \centering
  \setlength{\tabcolsep}{1.0mm}  
  \resizebox{1.0\linewidth}{!}{   
  \begin{tabular}{l|cc|cc|cc}
    \toprule
    \multicolumn{1}{c|}{\multirow{2}{*}{\textsc{Method}}} & \multicolumn{2}{c|}{KINS} & \multicolumn{2}{c|}{COCOA} & \multicolumn{2}{c}{COCOA-cls} \\
    & $\text{mIoU}_{f}$ & $\text{mIoU}_{o}$ & $\text{mIoU}_{f}$ & $\text{mIoU}_{o}$ & $\text{mIoU}_{f}$ & $\text{mIoU}_{o}$ \\
    \midrule
    PCNet\cite{pcnet} &78.02 &38.14 &76.91 &20.34 &-- &-- \\  
    VRSP\cite{vrsp} &80.70 &47.33 &78.98 &22.92  &79.93 &26.72 \\
    AISformer\cite{tran2024aisformeramodalinstancesegmentation} &81.53 &48.54 &72.69 &13.75 & -- & -- \\
    C2F-Seg\cite{gao2023coarsetofineamodalsegmentationshape} &82.22 &53.60 &80.28 &27.71  &81.71  &36.70 \\
    PLUG\cite{plug} &88.10 &61.42 &83.23 &32.88  &--  &-- \\
    \midrule
    $\textbf{Amodal SAM}$ &88.79 &63.12 &84.27 &59.94 &87.65 &54.34  \\
    \bottomrule
  \end{tabular}}
  \label{tab:close-domain}
\end{table}

\subsection{Evaluation of Amodal Image Segmentation}
\label{sec:evaluations}
\noindent\textbf{Closed-domain amodal segmentation.}
To have a comparison in the closed-domain scenario, \textit{i.e.}, the training and testing samples are from the same dataset, we evaluate the proposed methods with four representative works on the KINS, COCOA, and COCOA-cls datasets. This comparison is presented in Table \ref{tab:close-domain}, where ''Amodal SAM'' shows the performance when the proposed Amodal SAM is evaluated in the closed domain setting.

\begin{figure*}[]
    \centering
    \includegraphics[width=0.8\linewidth]{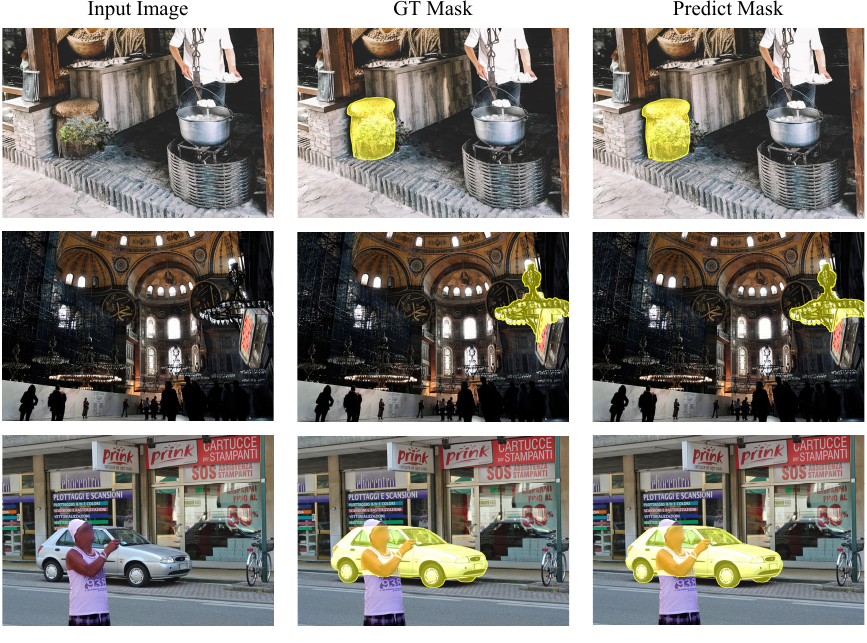}
    \caption{More qualitative results of Amodal SAM. "GT" represents the ground truth and the "Predict Mask" is the result of Amodal SAM prediction.}
    \label{fig:visible1}
\end{figure*}

\mypara{Open-world amodal segmentation.}
\label{sec:compare}
To further validate the cross-dataset robustness and zero-shot transferability of our proposed framework, we benchmark Amodal SAM against two state-of-the-art paradigms: the gestalt-principled pix2gestalt~\cite{pix} and the recently proposed SAMBA~\cite{SAMBA}. Unlike conventional evaluations, this zero-shot setting requires the models to generalize to unseen object categories and disparate environmental distributions without any fine-tuning.

We utilize the COCOA-cls and D2SA datasets as the primary evaluation testbeds, as they offer a diverse array of complex occlusion scenarios and high-density semantic classes. As quantitatively detailed in Table~\ref{tab:compare_methods}, our Amodal SAM significantly transcends existing methodologies. Specifically, Amodal SAM exhibits a substantial performance gain over SAMBA, demonstrating that our specialized Spatial Completion Adapter (SCA) and occlusion-aware optimization strategies are more effective at hallucinating the geometry of sequestered regions than existing adaptation schemes. These results underscore the superior generalization capability of our model in navigating the challenges of open-world amodal segmentation.

\begin{table}[h]
\caption{Comparison with Existing Methods}
\centering
\resizebox{0.7\linewidth}{!}{
\begin{tabular}{lcc}
    \toprule
    Method   &COCOA-cls & D2SA \\ 
    \midrule
    $\text{pix2gestalt}$ &79.08  & 81.82 \\
    $\text{SAMBA}$     &81.82  &90.87 \\
    \midrule
    $\text{Amodal SAM}$  &83.18  &91.62  \\
    \bottomrule
\end{tabular}}
\label{tab:compare_methods}
\end{table}

\mypara{Qualitative evaluation.}
The visual comparison is demonstrated in Fig. \ref{fig:visualization1}, from which we can observe that the proposed model shows superior amodal segmentation ability compared to the recent state-of-the-art methods.

\subsection{Evaluation of Amodal Video Segmentation}
\label{sec:evaluation_amodal_video_segmentatino}For comparison in the video amodal segmentation, we conducted additional experiments using our model for the video amodal segmentation task. Table \ref{tab:compare_video} presents a comparison between our approach and several competing methods on the FISHBOWL and MOViD-A datasets. Our Amodal SAM outperforms all current state-of-the-art techniques, including image-level and video-level baselines, across both datasets.

\begin{table}[h]
\caption{Performance comparison on video amodal segmentation datasets.}
\centering
\resizebox{0.9\linewidth}{!}{
\begin{tabular}{l|cc|cc}
\toprule
\multirow{2}{*}{Method}   & \multicolumn{2}{c}{FISHBOWL} & \multicolumn{2}{|c}{MOViD-A} \\ 
&  mIoU$_{f}$ &  mIoU$_{o}$ &  mIoU$_{f}$ &  mIoU$_{o}$ \\ \midrule
PCNET~\cite{pcnet}   &87.04   &65.02   &64.35   &27.31 \\
AISformer~\cite{tran2024aisformeramodalinstancesegmentation}   &-   &-  &67.72   &33.65 \\
SaVos~\cite{savos}   &88.63   &71.55   &60.01   &22.64 \\
C2F-Seg\cite{gao2023coarsetofineamodalsegmentationshape}   &91.68   &81.21   &71.67   &36.13 \\
\midrule
$\text{Amodal SAM}$     &92.74   &83.36   &73.06  &39.21 \\
\bottomrule
\end{tabular}}
\label{tab:compare_video}
\end{table}

\subsection{Ablation Study}
\label{sec:ablation_study}
This section shows the ablation study conducted to assess the impact of our designs. In all experiments, the model is trained on our customized amodal dataset obtained from SA-1B, and evaluated on amodal segmentation datasets, as depicted in Table \ref{tab:ablation}, Table \ref{tab:ablation_op} and Table \ref{tab:ablation_num}.

\mypara{Effect of the encoder-focused adaptation}. To illustrate that adjusting the image encoder results in enhanced generalization compared to adapting the mask decoder, we conduct experiments on the KINS~\cite{qi2019amodal} and COCOA-cls~\cite{follmann2018learninginvisibleendtoendtrainable} datasets for comparison. The results, as indicated in the 1st and 3rd rows of Table \ref{tab:ablation}, demonstrate that Encoder-focused Adaptation yields superior performance.

\mypara{Effect of the Spatial Completion Adapter (SCA)}. 
In contrast to standard convolution, the gated convolution adopted in the proposed SCA employs input region-specific masks to improve feature processing through a learned dynamic selection mechanism. To assess the SCA's effects in reconstructing occluded regions of target objects, we have a comparison with the adapter implemented with the conventional convolution layer, instead of the gated convolution. By comparing the results of ``EA+SCA'' and ``EA'' of Table \ref{tab:ablation}, we can observe that SCA yields much better results than the standard adapter. 

\begin{table}[h]
\caption{Ablation study of the proposed designs. ``EA'' denotes the Encoder-focused adaptation, and ``SCA'' denotes the spatial completion adapter. The 2nd row without ``SCA'' is implemented with the conventional adapter. The 3rd row without ``EA'' refers to the results obtained by inserting the adapters into the decoder. }

\centering
\resizebox{0.82\linewidth}{!}{
        \begin{tabular}{c|cc|cc}
            \toprule
            \multirow{2}{*}{Method} & \multicolumn{2}{c|}{KINS} & \multicolumn{2}{c}{COCOA-cls} \\ 
            & mIoU$_{f}$ &  mIoU$_{o}$ &  mIoU$_{f}$  & mIoU$_{o}$ \\ 
            \midrule
            EA + SCA  &85.43  &59.27 &83.18 &35.62 \\ 
            EA  &59.97 &30.63 &57.41 &31.47 \\
            SCA &65.04  &29.85 &58.18 &27.64 \\ 
            \bottomrule
        \end{tabular}}
        \label{tab:ablation}
\vspace{-1.2em}
\end{table}

\mypara{Ablation study on the optimization adaptation}
\label{sec:ablation_op}
In our optimization adaptation, we formulate a composite learning objective $\mathcal{L}$, encompassing not only the conventional $\mathcal{L}_{seg}$ but also regional consistency ($\mathcal{L}_\text{reg}$) and holistic topological regularization ($\mathcal{L}_\text{hol}$). The effectiveness of these newly introduced learning objectives, \textit{i.e.},  $\mathcal{L}_\text{reg}$ and $\mathcal{L}_\text{hol}$, can be validated through the results presented in Table~\ref{tab:ablation_op}.

\vspace{5pt}
\begin{table}[h]
\centering
\caption{Performance under different training objectives.}
\resizebox{0.9\linewidth}{!}{
\begin{tabular}{cc|cc|cc}
\toprule
\multicolumn{2}{c}{Method} & \multicolumn{2}{|c}{KINS} & \multicolumn{2}{|c}{TAOS Dataset} \\ 
$\mathcal{L}_\text{hol}$  & $\mathcal{L}_\text{reg}$ & mIoU$_{f}$ &  mIoU$_{o}$ &  mIoU$_{f}$  & mIoU$_{o}$ \\ \midrule
\checkmark  &\checkmark &85.43  &59.27 &90.85 &54.64 \\ 
$\times$  &\checkmark &83.05  &56.75 &89.04 &54.39 \\ 
\checkmark &$\times$ &82.79 &53.94  &89.97 &53.52  \\
$\times$ &$\times$  &81.02  &53.73 &88.45 &53.41 \\
\bottomrule
\end{tabular}}
\label{tab:ablation_op}
\end{table}

\mypara{Ablation study on the number of adapters}. As detailed in Section~\ref{sec:model_adaptation}, we integrated three SCAs into the image encoder. To assess the impact of varying the number of SCAs on the performance of Amodal SAM, we conducted experiments using different numbers of SCAs on the KINS~\cite{qi2019amodal} and COCOA-cls~\cite{follmann2018learninginvisibleendtoendtrainable} datasets, with the results summarized in Table~\ref{tab:ablation_num}. The results demonstrate that our method of integrating SCAs achieves effective performance while minimally increasing the model's parameters, thereby maintaining efficient inference speed. 

\begin{figure*}[t]
    \centering
    \includegraphics[width=\linewidth]{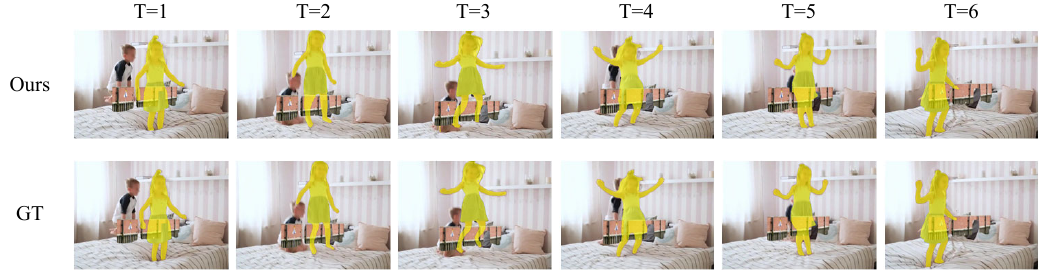}
    \caption{This figure illustrates the viability of transitioning our method to SAM-2.}
    \label{fig:visualization2}
\end{figure*}

\vspace{10pt}
\begin{table}[h]
\caption{Ablation study on the number of adapters}
\centering
\resizebox{0.9\linewidth}{!}{
\begin{tabular}{l|cc|cc}
\toprule
\multirow{2}{*}{Method}   & \multicolumn{2}{c}{KINS} & \multicolumn{2}{|c}{COCOA-cls} \\ 
&  mIoU$_{f}$ &  mIoU$_{o}$ &  mIoU$_{fu}$ &  mIoU$_{o}$ \\ \midrule
one adapter         &78.05  &51.92  &79.94 &29.58 \\
two adapter     &83.26  &56.34 &81.74 &32.97  \\
full model        &85.43  &59.27  &83.18 &35.62 \\
\bottomrule
\end{tabular}}
\label{tab:ablation_num}
\end{table}

\mypara{Effectiveness of use points as prompt}
\label{sec:prompt}
SAM~\cite{kirillov2023segment} features a versatile prompt encoder capable of handling various types of input prompts, such as points, boxes, and text. In the previous experiments, we used boxes as the prompt input. To evaluate whether Amodal SAM can also perform amodal segmentation with points as prompts, we designed this experiment. For comparison with the prior use of boxes, we randomly sampled two points within the visible region of the target object as the prompt input during both training and inference. 

The experiments were conducted on the KINS~\cite{qi2019amodal} and COCOA~\cite{zhu2016semanticamodalsegmentation} dataset, and the results are presented in Table \ref{tab:compare_prompt}. Experimental results show that amodal sam still achieves good performance using points as the prompt input. For the COCOA~\cite{zhu2016semanticamodalsegmentation} dataset, where object occlusion scenarios are more diverse and complex, using boxes as prompts may introduce ambiguity. In contrast, using points as prompts can mitigate this issue and even yield better performance.

\begin{table}[h]
\caption{Performance comparison of different prompts.}
\centering
\resizebox{0.9\linewidth}{!}{
\begin{tabular}{l|cc|cc}
\toprule
\multirow{2}{*}{Prompt}   & \multicolumn{2}{c}{KINS} & \multicolumn{2}{|c}{COCOA-cls} \\ 
&  mIoU$_{f}$ &  mIoU$_{o}$ &  mIoU$_{f}$ &  mIoU$_{o}$ \\ \midrule
$\text{Box}$        &85.43  &59.27 &83.18 &35.62 \\
$\text{Points}$     &84.74  &58.52 &82.76 &36.28  \\
\bottomrule
\end{tabular}}
\label{tab:compare_prompt}
\end{table}

\section{Concluding Remarks}
\mypara{Summary.}
In this work, we presented Amodal SAM, a framework that successfully extends SAM's capabilities to open-world amodal segmentation through three synergistic aspects: model, data, and optimization. Our approach achieves state-of-the-art performance across multiple benchmarks while maintaining zero-shot capabilities, and can be readily extended to video applications. We believe this work can provide a solid foundation for future research in open-world visual scene understanding.

\mypara{Limitations \& Future Work.}
The introduced TAOS serves as an autonomous data generation pipeline, eliminating the need for manual intervention. While a small portion of authentic occlusions may remain unprocessed due to missing annotations, this limitation does not significantly impact the method's effectiveness. Notably, TAOS-generated data consistently improves performance across benchmark datasets, and addressing this constraint presents a promising opportunity to further enhance amodal segmentation capabilities.

\bibliographystyle{IEEEtran}
\bibliography{IEEEabrv,mainbib}

\end{document}